\documentclass[11pt, letterpaper]{article}
\usepackage{amsmath}
\usepackage{amssymb}
\usepackage{amsthm}
\usepackage{booktabs}
\usepackage{graphicx}
\usepackage{hyperref}
\usepackage{url}
\usepackage{algorithm}
\usepackage{algorithmicx}
\usepackage{algpseudocode}
\usepackage{caption}
\usepackage{subcaption}
\usepackage{booktabs}
\usepackage{authblk}
\usepackage{appendix}
\usepackage[utf8]{inputenc}
\usepackage[T1]{fontenc}
\usepackage[english]{babel}
\usepackage{geometry}
\usepackage{xcolor}
\usepackage{fancyhdr}
\usepackage{microtype}
\usepackage{parskip}
\usepackage{tabularx}
\usepackage{enumitem}

\hypersetup{
    colorlinks=true,
    linkcolor=blue,
    filecolor=magenta,      
    urlcolor=cyan,
    citecolor=red,
}

\definecolor{headercolor}{RGB}{0, 50, 100}
\title{PharmacyGPT: the Artificial Intelligence Pharmacist and an Exploration of AI for ICU Pharmacotherapy Management}
\newcommand*\samethanks[1][\value{footnote}]{\footnotemark[#1]}
\author[1]{Zhengliang Liu \thanks{Co-first authors.}}
\author[1]{Zihao Wu \samethanks}
\author[1]{Mengxuan Hu \samethanks}
\author[1]{Shaochen Xu \samethanks}
\author[2]{Bokai Zhao}
\author[1]{Lin Zhao}
\author[3]{Tianyi Zhang}
\author[1]{Haixing Dai}
\author[1]{Yiwei Li}
\author[3]{Xianyan Chen}
\author[2]{Ye Shen}
\author[4]{Sheng Li}
\author[5]{Quanzheng Li}
\author[5]{Xiang Li}
\author[6]{Brian Murray}
\author[1]{Tianming Liu}
\author[7]{Andrea Sikora}

\affil[1]{School of Computing, University of Georgia}
\affil[2]{Department of Epidemiology \& Biostatistics, University of Georgia}
\affil[3]{Department of Statistics, University of Georgia}
\affil[4]{School of Data Science, University of Virginia}
\affil[5]{Massachusetts General Hospital and Harvard Medical School}
\affil[6]{Department of Pharmacy, University of North Carolina Medical Center}
\affil[7]{Department of Clinical and Administrative Pharmacy, University of Georgia College of Pharmacy}

\date{}

\begin{document}

\maketitle

\begin{abstract}
In this study, we introduce PharmacyGPT, a novel framework to assess the capabilities of large language models (LLMs) such as ChatGPT and GPT-4 in emulating the role of clinical pharmacists. Our methodology encompasses the use of LLMs to generate comprehensible patient clusters, formulate medication plans, and forecast patient outcomes. We conducted our investigation using real data acquired from the intensive care unit (ICU) at the University of North Carolina Health System (UNCHS). Our analysis offers  insights into the potential applications and limitations of LLMs in the field of clinical pharmacy, with implications for both patient care and the development of future AI-driven healthcare solutions. By evaluating the performance of PharmacyGPT, we aim to contribute to the ongoing discourse surrounding the integration of artificial intelligence in healthcare settings, ultimately promoting the responsible and efficacious use of such technologies.
\end{abstract}

\section{Introduction}
\label{sec:introduction}

In recent years, the development of large language models (LLMs) has shown remarkable potential in a multitude of applications across various domains \cite{zhao2023brain,holmes2023evaluating,liu2023summary,li2023artificial}. Despite their impressive generalization capabilities, the performance of LLMs in specific fields (e.g., pharmacy), remains under-investigated and holds untapped potential. Here, we present PharmacyGPT, the first study to apply LLMs, specifically ChatGPT and GPT-4, to a range of clinically significant problems in the realm of comprehensive medication management in the intensive care unit (ICU) \cite{sikora2023evaluation,sikora2022impact,sikora2023pharmacophenotype}.

PharmacyGPT aims to explore the capabilities of LLMs in addressing diverse problems of clinical importance \cite{snoswell2023pharmacist}, such as patient outcome prediction, AI-based medication decisions, and interpretable clustering analysis of patients. These applications hold promise for improving patient care \cite{zhu2023can} by empowering a comprehensive suite of AI-enhanced clinical decision support tools for clinical pharmacists.

Due to the degree of domain-specific knowledge in fields such as clinical pharmacy, LLMs require domain-specific engineering. Here, a combination of dynamic prompting and iterative optimization was applied. To enhance the performance of LLMs in the pharmacy domain, we developed a dynamic prompting approach that leverages the in-context learning capability \cite{brown2020language} of LLMs by constructing dynamic contexts using domain-specific data, which is beneficial for adapting language models to specialized fields \cite{gu2021domain,liu2023radiology,rezayi2022agribert,rezayi2022clinicalradiobert,cai2022coarse}. This novel method enables the model to acquire contextual knowledge from semantically similar examples in existing data \cite{brown2020language,ma2023impressiongpt} thereby improving its performance in specialized applications.

Additionally, we designed an iterative optimization algorithm that performs automatic evaluation on the generated prescription results and composes the corresponding instruction prompts to further optimize the model. This  technique allows the refinement of PharmacyGPT without requiring additional training data or fine-tuning the LLMs, paving the way for efficient adaptation of LLMs in other specialized fields.

\begin{figure*}[hbt!]
	\centering
	\includegraphics[width=1\textwidth]{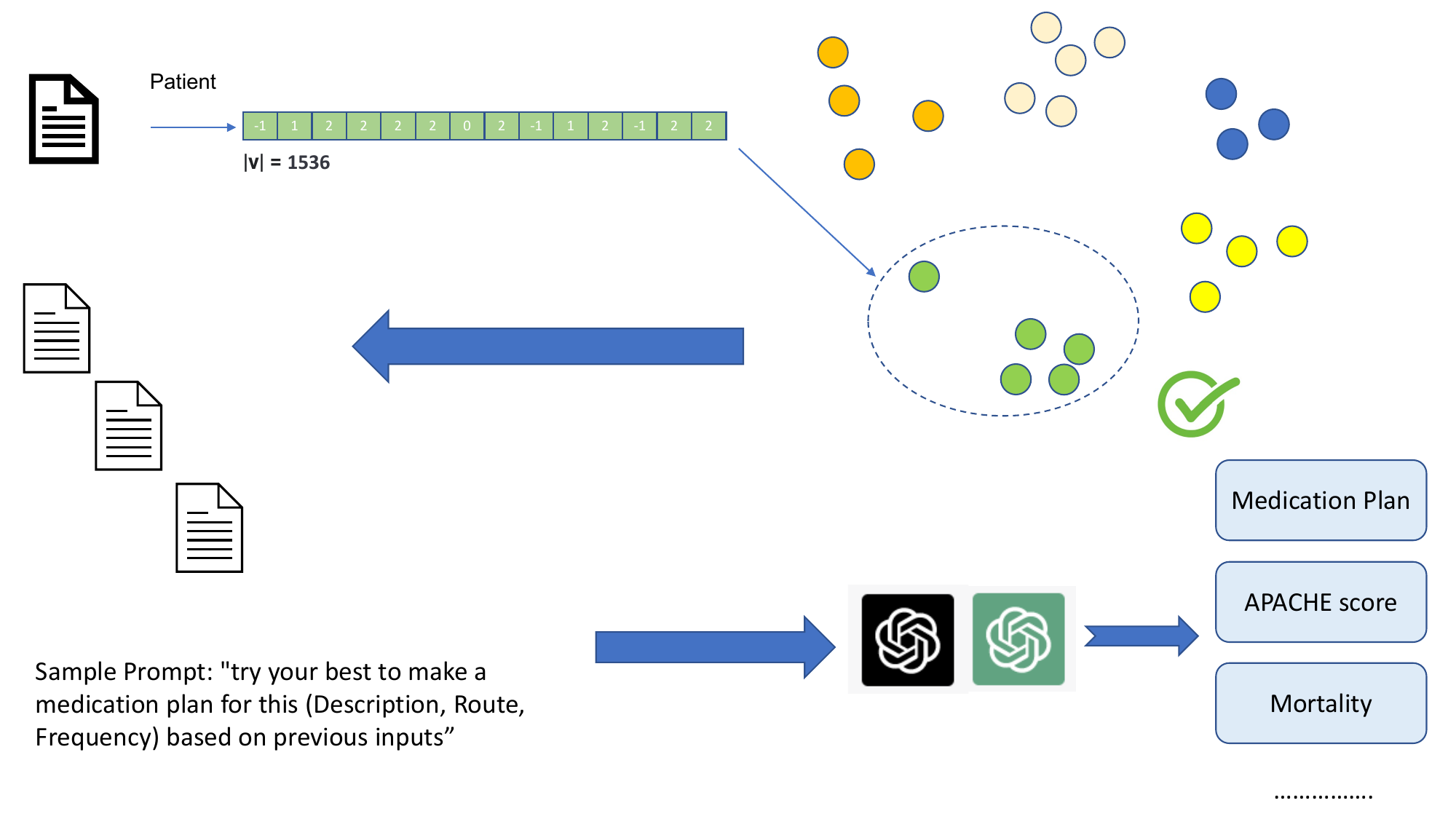} 
	\caption{Generating context for ChatGPT and GPT-4}
	\label{fig3}
\end{figure*}

We compared various approaches to optimally use LLMs for a range of pharmacy tasks and aimed to establish a use case portfolio to inspire future research exploring LLM application to pharmacotherapy-related questions.

This work has several goals: 
\begin{itemize}
  \item 1) To open new dialogues and stimulate discussion surrounding the application of LLMs in pharmacy. 
  \item 2) To provide directions for investigators for future improvements in LLM-based pharmacy applications.
  \item 3) To evaluate the strengths and limitations of ChatGPT and GPT-4 in the pharmacy domain. 
  \item 4) To provide insights for tailored data collection strategies to maximize the potential of LLMs in pharmacy and other specialized fields.
  \item 5) To present an effective framework and use case collection for applying LLM to pharmacy. 
\end{itemize}

In conclusion, this paper introduces PharmacyGPT, a groundbreaking work that applies LLMs to a range of clinically significant problems in pharmacy. By establishing a foundation for future research and development, PharmacyGPT has the potential to revolutionize pharmacy practices and enhance the overall quality of healthcare services while addressing its key motivations and goals, ultimately contributing to a deeper understanding and more effective use of LLMs in specialized domains.

\section{Related Work}
\label{sec:relatedwork}

\subsection{Large language models in healthcare}
Transformer-based language models such as Bidirectional Encoder Representations from Transformer (BERT) \cite{devlin2018bert} and the Generative Pre-trained Transformer (GPT) series \cite{radford2018improving,radford2019language,brown2020language}, have revolutionized the field of natural language processing (NLP) by outperforming earlier approaches like Recurrent Neural Network (RNN)-based models \cite{fan2023bibliometric,holmes2023evaluating,zhao2023brain,liu2022survey} in a variety of tasks. Existing transformer-based language models can be broadly classified into three categories: masked language models (e.g., BERT), autoregressive models (e.g., GPT-3), and encoder-decoder models (e.g., Bidirectional Auto-Regressive Transformer or BART \cite{lewis2019bart}). The recent development of very large language models (LLMs) grounded in the transformer architecture but built on a much grander scale, including ChatGPT and GPT-4 \cite{liu2023summary}, Bloomz \cite{muennighoff2022crosslingual}, LLAMA \cite{touvron2023llama} and Med-PaLM 2 \cite{singhal2022large} has gained momentum.

The primary objective of LLMs is to accurately learn context-specific and domain-specific latent feature representations from input text \cite{devlin2018bert,liu2023summary,liu2023context,zhou2023fine,cai2023exploring,liao2023mask}. For example, the vector representation of the word "prescription" could vary significantly between the pharmacy domain and general usage. Smaller language models such as BERT often necessitate pre-training and supervised fine-tuning on downstream tasks to achieve satisfactory performance. However, LLMs typically do not require further fine-tuning and model updates but still deliver competitive performance on diverse downstream applications \cite{dai2023chataug,cai2023exploring,liao2023differentiate,dai2023ad,zhong2023chatabl}.

LLMs like ChatGPT and GPT-4 have significant potential in healthcare applications due to their advanced natural language understanding (NLU) capabilities. The massive amounts of text and other data generated in clinical practices can be leveraged through LLM-powered tools. LLMs can handle diverse and complex tasks like clinical triage classification \cite{wang2021reduce}, medical question-answering \cite{wu2023exploring}, HIPAA-compliant data anonymization \cite{liu2023deid}, radiology report summarization \cite{ma2023impressiongpt}, clinical information extraction \cite{chen2020joint} or dementia detection \cite{saltz2021dementia}. In addition, the Reinforcement Learning from Human Feedback (RLHF) \cite{ouyang2022training} process incorporates human preferences and values into ChatGPT and its successor, GPT-4, making it particularly suitable for understanding patient-centered guidelines and human values in healthcare applications.

We aim to present the first study employing LLMs for a wide range of tasks and problems of interest to pharmacy practice in the intensive care unit (ICU). 

\subsection{Reasoning with LLMs}
LLMs have shown great potential in high-level reasoning abilities. For example, GPT-3 has demonstrated common-sense reasoning through in-context learning, and Wei et al. \cite{wei2022chain} found that LLMs can perform better in arithmetic, deductive, and common-sense reasoning when given carefully prepared sequential prompts, decomposing multi-step problems.

\textbf{Deductive reasoning:} Wu et al. \cite{wu2023exploring} compared the deductive reasoning abilities of ChatGPT and GPT-4 in the specialized domain of radiology, revealing that GPT-4 outperforms ChatGPT, with fine-tuned models requiring significant data to match GPT-4's performance. The results of this investigation suggest that creating generic reasoning models based on LLMs for diverse tasks across various domains is viable and practical.

Ma et al. explored LLMs' \cite{ma2023impressiongpt} ability to comprehend radiology reports using a dynamic prompting paradigm. They developed an iterative optimization algorithm for automatic evaluation and prompt composition, achieving state-of-the-art performance on MIMIC-CXR and OpenI datasets without additional training data or LLM fine-tuning.

\textbf{Abductive reasoning:} Jung et al. \cite{jung2022maieutic} improved LLMs' ability to make logical explanations through maieutic prompting, but this study was limited to question-answering style problems. Zhong et al. \cite{zhong2023chatabl} improved from the well-known ABL framework \cite{zhou2019abductive} and proposed the first comprehensive abductive learning pipeline based on LLMs. 

\textbf{Chain-of-thought:} Chain-of-thought reasoning (CoT) is a problem-solving approach that breaks complex problems into smaller, manageable steps, resembling how humans naturally solve complicated problems \cite{saparov2022language}. In the context of LLMs, CoT aims to improve the model's accuracy and coherence by encouraging step-by-step reasoning. A zero-shot approach asking the LLM to "think step by step" \cite{kojima2022large} significantly improves reasoning performance across benchmarks. Providing LLMs with more examples (few-shot context learning) \cite{wei2022chain} enhances their performance through in-context learning. Overall, CoT enables LLMs to effectively tackle complex reasoning tasks.

\section{Methods}
\label{sec:methods}

\subsection{Data description}

This dataset was based on a study cohort of 1,000 adult patients admitted for at least 24 hours to a medical, surgical, neurosciences, cardiac, or burn ICU at the University of North Carolina Health System (UNCHS) between October 2015 and October 2020. Only the patient's first ICU admission was included in the analysis. The data were extracted from the UNCHS Electronic Health Record (EHR) housed in the Carolina Data Warehouse (CDW) by a trained in-house data analyst.

The dataset contains patient demographics, medication administration record (MAR) information, and patient outcomes, such as common ICU complications. Demographics include age, sex, admission diagnosis, ICU type, Medication Regimen Complexity-Intensive Care Unit (MRC-ICU) \cite{Sikora2023-px, webb2022descriptive, Webb2022-yb, Sikora2022-wa, Newsome2019-jy, Newsome2020-wf, Olney2022-ed, Smith2022-ez, Gwynn2019-ui, Al-Mamun2021-xv, Sikora2023-fg}, score at 24 hours, and APACHE II score \cite{larvin1989apache} at 24 hours. The MRC-ICU \cite{newsome2020medication,newsome2019characterization} is a validated score summarizing the complexity of prescribed medications in the ICU \cite{newsome2020multicenter,olney2022medication}. MAR information consists of drug, dose, route, duration, and timing of administration \cite{smith2022medication,gwynn2019development,al2021development,}.

Patient outcomes included mortality, ICU length of stay, delirium occurrence (defined by a positive CAM-ICU score), duration of mechanical ventilation, duration of vasopressor use, and acute kidney injury (defined by the presence of renal replacement therapy or a serum creatinine greater than 1.5 times baseline). Other than textual descriptions of patient data, a binary value of 1 was assigned to indicate they received a specific drug order, including drug, dose, strength, and formulation/route. Categorical features for patient outcomes were relabeled as numeric values, and any unknown or missing entities were counted as absences from that event.

\subsection{Creating Interpretable Patient Clusters}
Here, we describe the process for generating interpretable patient clusters using LLM embeddings and hierarchical clustering. The following pseudo-code algorithm outlines the steps:

\begin{enumerate}
    \item Feed patient information (i.e., age, sex, diagnosis, and ICD10 problem list) into GPT-3 to generate embedding vectors of size 1536.
    \item Use the generated embeddings as input for hierarchical clustering to create patient clusters.
\end{enumerate}

\begin{algorithm}[hbt!]
\caption{Interpretable Patient Clustering}\label{alg:patient_clustering}
\begin{algorithmic}[1]
\Procedure{CreateClusters}{$patient\_info$}
    \State $embedding\_vectors \gets \textit{generate\_gpt3\_embeddings}(patient\_info)$
    \State $patient\_clusters \gets \textit{apply\_hierarchical\_clustering}(embedding\_vectors)$
    \State \textbf{return} $patient\_clusters$
\EndProcedure
\Function{generate\_gpt3\_embeddings}{$patient\_info$}
    \State $embeddings \gets []$
    \For{each $p$ in $patient\_info$}
        \State $embedding \gets \textit{GPT-3\_embedding}(p)$
        \State $\textit{append}(embeddings, embedding)$
    \EndFor
    \State \textbf{return} $embeddings$
\EndFunction
\Function{apply\_hierarchical\_clustering}{$embedding\_vectors$}
    \State $clusters \gets \textit{hierarchical\_clustering}(embedding\_vectors)$
    \State \textbf{return} $clusters$
\EndFunction
\end{algorithmic}
\end{algorithm}

The above algorithm describes the process of generating interpretable patient clusters using ChatGPT embeddings and hierarchical clustering. In this method, patient information is first transformed into 1536-dimensional embeddings using GPT-3 \cite{brown2020language,openaiImprovedEmbedding}. These embeddings are then used as input for a hierarchical clustering algorithm, which groups patients based on the similarity of their embeddings. This approach aims to create accurate and interpretable patient clusters for use in clinical decisions.


\subsection{Iterative Optimization}

In the iterative optimization process for PharmacyGPT, we aimed to enhance the performance of the model when generating various patient-related outputs, such as mortality, length of ICU stay, APACHE II score range, or medication plan at 24 hours. This optimization process is based on an iterative feedback loop, which adjusts the input prompts provided to the model based on its performance in previous iterations. 

\begin{algorithm}[hbt!]
    \caption{Iterative Optimization Algorithm for Patient Data}
    \label{algorithm_1}
    \begin{algorithmic}[1]
        \State \textbf{Input:} Patient portrait based on demographics and symptoms
        \State \textbf{Initialize:} $I$=Iteration times, $Thre_S$=Threshold of evaluation score, $iter$=0
        \State \textbf{Definition:} $GPT$ means ChatGPT, $Prompt_{Dy}$ means our dynamic prompt, $Prompt_{Iter}$ means our iterative prompt, $L$ means evaluate function, $M_{good}$ and $M_{bad}$ represent prompt merging with good and bad response
        \While {$iter < I$}
            \If{$iter=0$}
                \State{$output = GPT(Prompt_{Dy})$}
            \Else
                \State{$output = GPT(Prompt_{Iter})$}
            \EndIf
            \State{$score = \frac{1}{N_s} \sum\limits_{i=0}^{N_s} L(output, Impression_i) $}
            \If{$score > Thre_S$}
                \State{$Prompt_{Iter} = M_{good}(Prompt_{Dy}, output)$}
            \Else
                \State{$Prompt_{Iter} = M_{bad}(Prompt_{Dy}, output)$}
            \EndIf  
            \State{$iter++$}
        \EndWhile
    \end{algorithmic}
\end{algorithm}

\begin{figure*}[hbt!]
	\centering
	\includegraphics[width=1\textwidth]{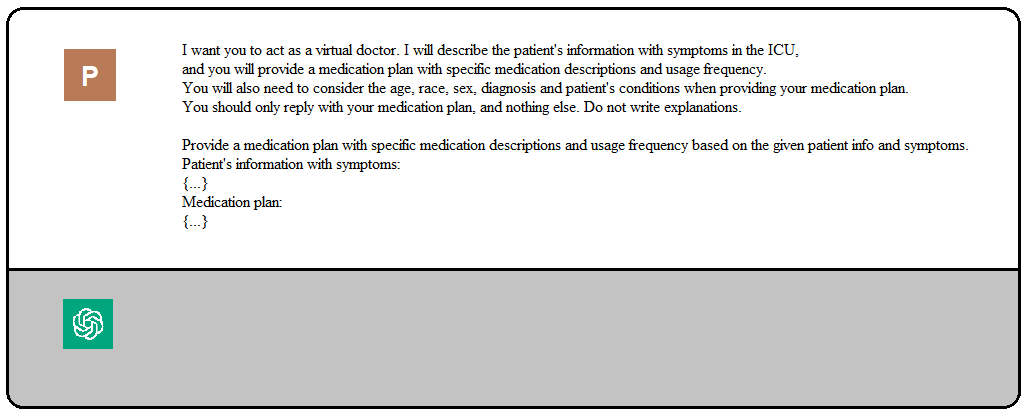} 
	\caption{Initial prompt used for medication plan generation}
	\label{fig7}
\end{figure*}

\begin{figure*}[hbt!]
	\centering
	\includegraphics[width=1\textwidth]{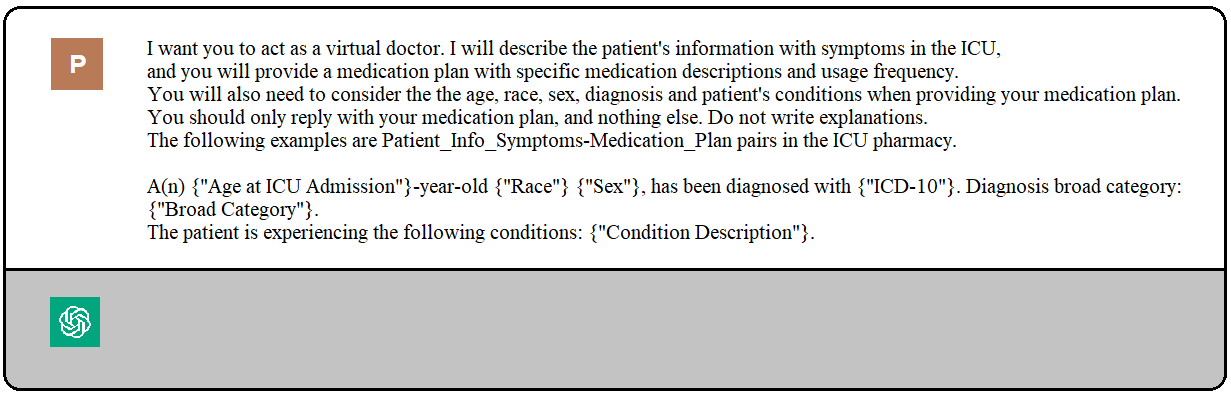} 
	\caption{Final prompt used for medication plan generation}
	\label{fig8}
\end{figure*}

Initially, we inputted a dynamic prompt containing a patient portrait based on demographics and symptoms to the ChatGPT model (see Figure \ref{fig7}). After this iteration, we calculated the ROUGE-1 score by comparing the model-generated output with the ground truth of the medications ordered in the first 24 hours. Based on the score, the input prompt was modified for the next iteration. If the score was above a predefined threshold, we merged the prompt with the generated output in a manner that encouraged the model to produce a similar response in the next iteration. Otherwise, we modified the prompt to discourage the model from producing the same type of output. The iterative process continued until a predetermined number of iterations was reached. Through this optimization method, PharmacyGPT learned to improve its predictions and recommendations over time, as it is continually guided by the evaluation score and feedback from previous iterations. This approach enables PharmacyGPT to adapt and refine its output based on a better understanding of the patient's condition and the desired outcomes. The final prompt used can be seen in Figure \ref{fig8}. 

\begin{figure*}[hbt!]
	\centering
	\includegraphics[width=1\textwidth]{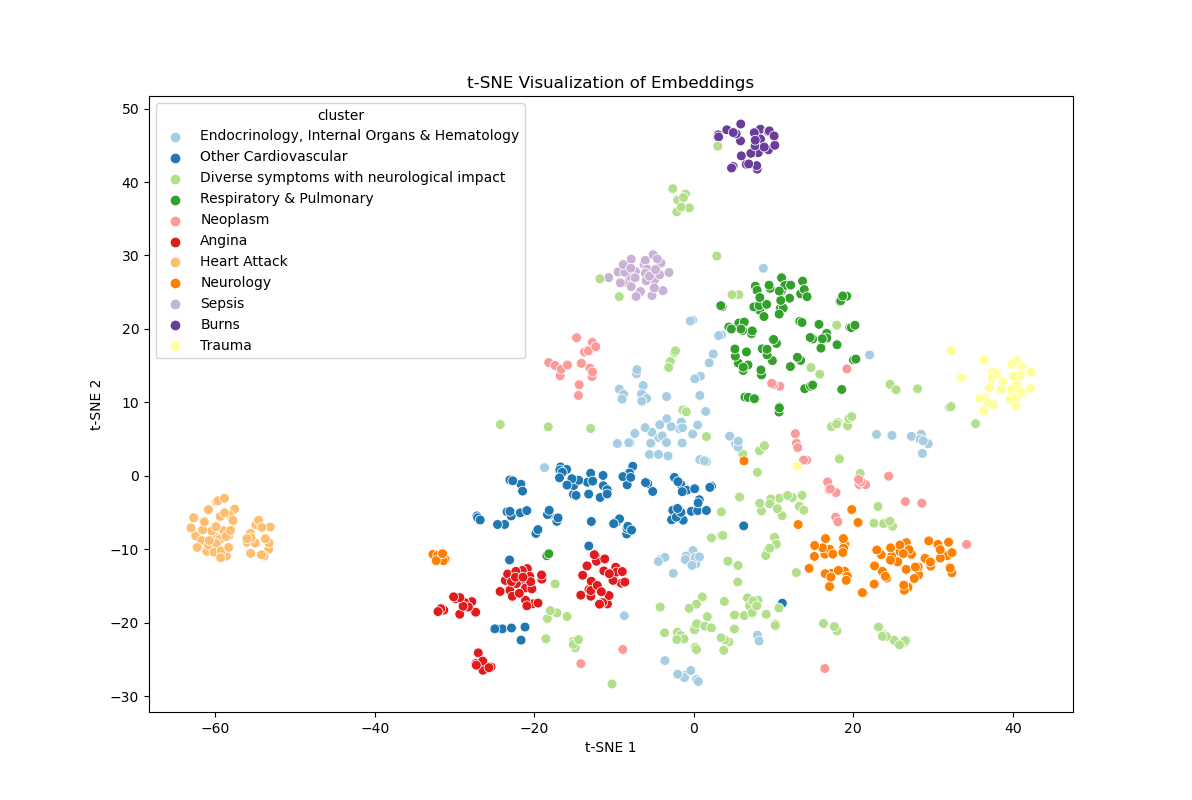} 
	\caption{Interpretable Clusters for real ICU data. Patient age, sex, icd-10 diagnosis, and ICD10 problem list were input to generate ChatGPT embeddings and hierarchical clustering as depicted above, which resulted in meaningful cluster separation. These patient clusters were created by embeddings generated by ADA-002 and clustered via agglomerative clustering techniques and visualized using T-SNE (t-distributed Stochastic Neighbor Embedding).}
	\label{fig1}
\end{figure*}

\section{Results}
\label{sec:results}

\subsection{Interpretable clustering}
Our clustering methodology yielded clusters that closely aligned with the ICD-10 code categories of patients, indicating a high level of interpretability (see Figure \ref{fig1}). This alignment demonstrates the effectiveness of our approach in generating meaningful and coherent patient groupings. 

\begin{figure*}[hbt!]
	\centering
	\includegraphics[width=1\textwidth]{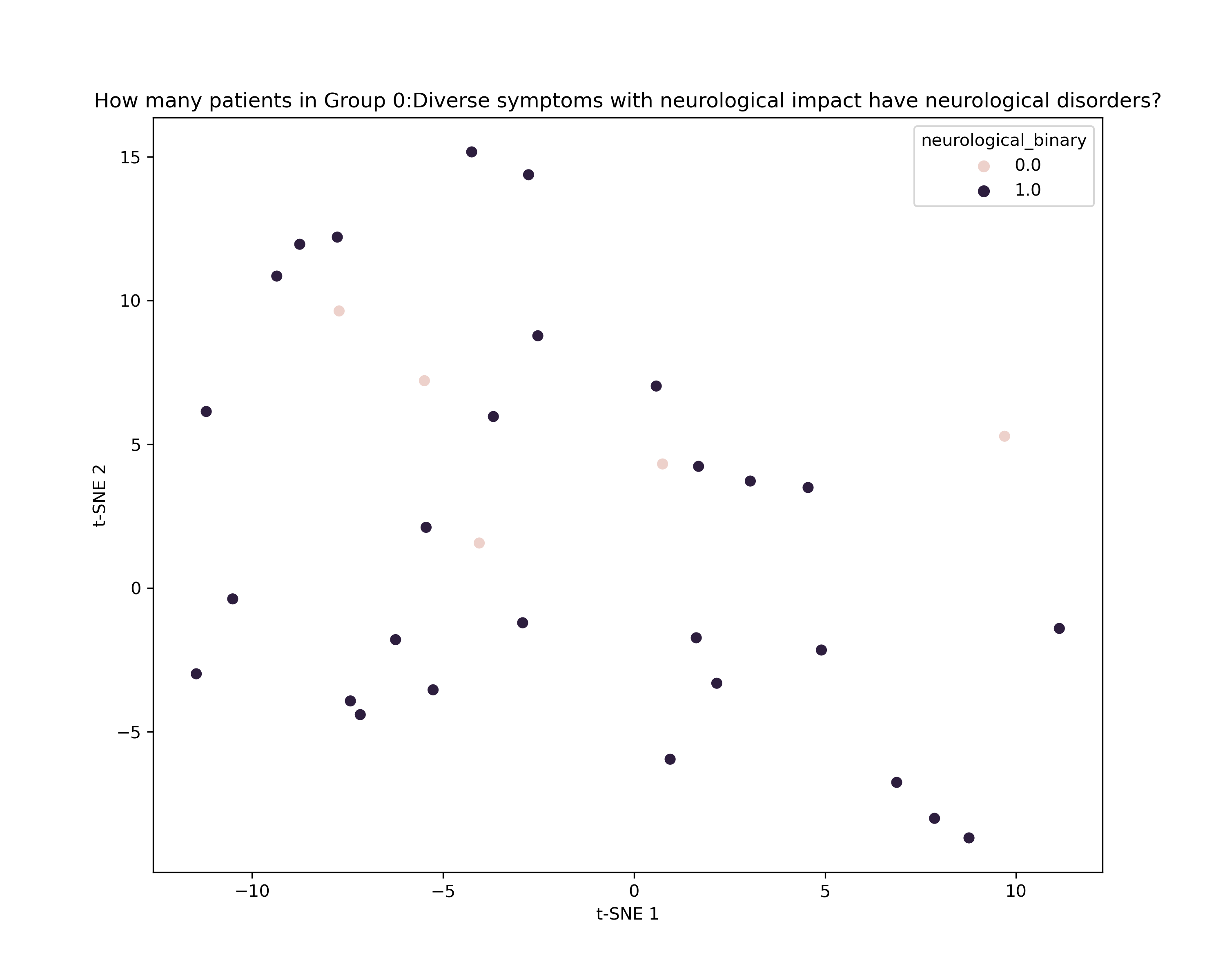} 
	\caption{Here, dots represent patients who were clustered to the neurology diagnosis group, with black dots showing a neurology associated ICD10 diagnosis. }
	\label{fig2}
\end{figure*}

The group labeled as "Diverse symptoms with neurological impact" is the most sparsely spread among the clusters, exhibiting a diverse range of symptoms on the surface. However, our embedding analysis reveals that the majority of patients in this group share a common underlying condition: neurological disorders. This observation suggests that despite the presence of various other symptoms, patients in this group are primarily affected by neurological issues (see Figure \ref{fig2}). Our approach successfully uncovered this underlying similarity, demonstrating the capability of our clustering method in identifying meaningful connections between patients with seemingly disparate symptom presentations.

\subsection{Predicting patient outcomes}

\textbf{Hospital Mortality}
We used ChatGPT to predict hospital mortality based on patient information including age, race, sex, ICD-10 admission diagnosis, and comorbidities). Table \ref{table:mortal} results indicate that imbalanced data significantly reduced precision and F1 scores, as the deceased class in the test set contained only 46 samples, while ChatGPT tended to make predictions in a balanced manner. Also, different prompts did not contribute much to improvement. Specifically, the last three rows attempted to align the diagnosis categories of the test set with those demonstrated in the prompts: Rand-5 randomly selects five demonstrations from the training set and incorporates them into the prompt; Freq-5 randomly selects one demonstration from the top five most frequent diagnosis categories; Bcat-rand-5 randomly chooses five samples with the same diagnosis category as the tested sample; and Sim-5 selects the five most similar patient information records as demonstrations.
\begin{table}[hbt!]
\centering
\caption{Mortality prediction}
\label{table:mortal}
\begin{tabular}{lcccc}
\toprule
\textbf{Model} & \textbf{Accuracy} & \textbf{Precision} & \textbf{Recall} & \textbf{F1 Score} \\
\midrule
rand\_5-shot      & 0.7549 & 0.3750 & 0.7095 & 0.4898 \\
freq\_5-shot      & 0.6455 & 0.1667 & 0.6364 & 0.2642 \\
bcat\_rand\_5-shot & 0.6602 & 0.2051 & 0.7647 & 0.4262 \\
sim\_5-shot       & 0.6699 & 0.2821 & 0.6471 & 0.3929 \\
\bottomrule
\end{tabular}
\end{table}

\textbf{APACHE II}
We employed ChatGPT and GPT-4 to predict APACHE II scores based on patient information (age, race, sex, ICD-10 admission diagnosis, and comorbidities). Table \ref{table:apache} results reveal that GPT-4 significantly enhanced accuracy. Accuracy is the ratio of correctly classified instances to the total number of instances. Precision focuses on the proportion of instances where the model predicts a positive outcome that actually turns out to be positive. It is a measure of the accuracy of positive class predictions. Recall is the model's ability to correctly identify positive classes. It is the proportion of correct identifications by the model from all actual positive examples. The F1 score is the harmonic mean of precision and recall, and it attempts to find a balance between the two.

\begin{table}[hbt!]
\centering
\caption{APACHE II score prediction}
\label{table:apache}
\begin{tabular}{lcccc}
\toprule
\textbf{Model} & \textbf{Accuracy}\\
\midrule
rand\_5-shot      & 0.1818 \\
freq\_5-shot      & 0.1091 \\
bcat\_rand\_5-shot & 0.2000 \\
sim\_5-shot       & 0.1636 \\
GPT-4\_rand\_5     & 0.3727 \\
GPT-4\_sim\_5      & 0.4364 \\
\bottomrule
\end{tabular}
\end{table}

\subsection{Prescribing medication plans}
We employed GPT-4 to generate medication plans for ICU patients using the same patient information as above. We then compared these plans to the medication plans at the 24 hour point used in practice. Result can be seen in Table \ref{tab:my_label}. ROUGE-1, ROUGE-2, and ROUGE-L are the three main ROUGE (Recall-Oriented Understudy for Gisting Evaluation) metrics used to evaluate automatic text summarization. ROUGE-1 focuses on the overlap of words (1-grams), and ROUGE-2 is based on the overlap of bigrams (2-grams). ROUGE-L focuses on the Longest Common Subsequence (LCS). This method does not require consecutive word sequence matching but rather identifies the longest sequence of sequential matching words between the candidate abstract and the reference abstract.

\begin{table}[h]
    \centering
    \begin{tabular}{lccc}
        \toprule
        & \textbf{Rouge1} & \textbf{Rouge2} & \textbf{Rouge3} \\
        \midrule
        Pharmacy\_GPT & 0.0739 & 0.0069 & 0.0519 \\
        LLAMA2 & 0.013 & 0.0001 & 0.0101 \\
        ChatGPT & 0.0404 & 0.0016 & 0.0259 \\
        GPT4 & 0.0466 & 0.0034 & 0.0287 \\
        \bottomrule
    \end{tabular}
    \caption{Rouge score of each model. ROUGE (Recall-Oriented Understudy for Gisting Evaluation) score is an evaluation metric widely used in the field of natural language processing, especially in text summarization and machine translation. It is mainly used to evaluate the quality of automatically generated summaries or translations. The core idea of ROUGE is to evaluate candidate abstracts by comparing them to a set of reference abstracts (usually human-written).}
    \label{tab:my_label}
\end{table}

\begin{figure*}[t]
	\centering
	\includegraphics[width=0.8\textwidth]{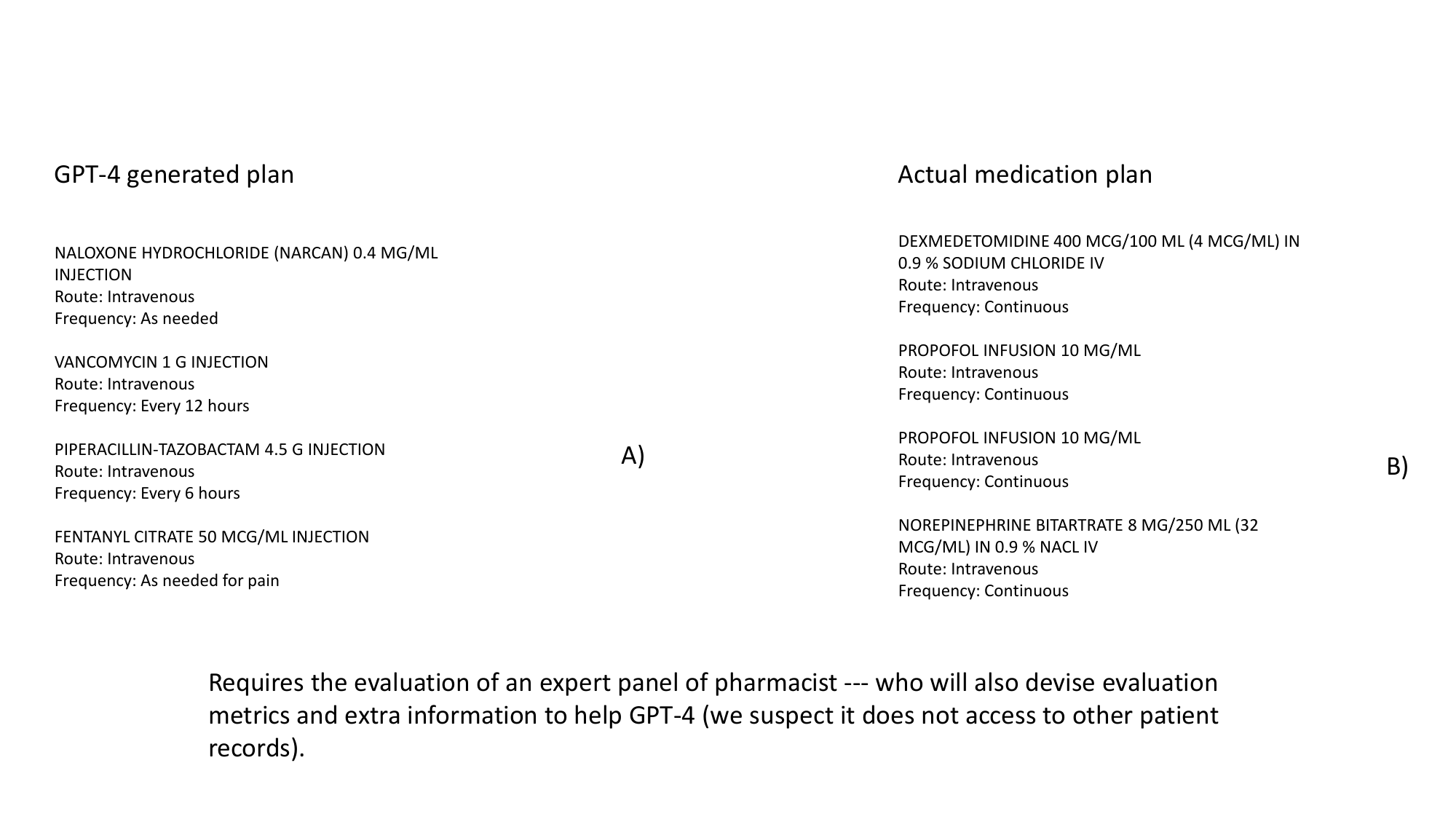} 
	\caption{Medication plan generated by GPT-4 v.s. the actual plan}
	\label{fig5}
\end{figure*}


\section{Discussion}
\label{sec:discussion}

We observed that the GPT-4 model fed with dynamic context and similar samples achieved the highest accuracy among all tested models. This demonstrates the potential of GPT-4 for highly specialized domains when applied with carefully designed context and sample selection. However, we found that the precision and recall scores were not particularly high for any of the approaches, which may be due to imbalances in the dataset (i.e., alive patients outnumbering deceased patients by a 9 to 1 ratio). This imbalance could have adversely impacted the performance of the models, particularly in terms of precision and recall. Future work could involve addressing this data imbalance and exploring other evaluation metrics better suited for imbalanced datasets to gain a more comprehensive understanding of model performance.

\subsection{Addressing AI anxiety in healthcare}
The issue of the 'black-box effect' and associated AI anxiety in healthcare \cite{sparrow2020high} must be addressed \cite{oh2019physician}. It is crucial to recognize the potential benefits of AI, such as improved patient care quality and increased efficiency in clinical practice, while also acknowledging the limitations of current AI systems, specifically large language models (LLMs) like ChatGPT and GPT-4 \cite{li2023artificial,holmes2023evaluating}.

Our study has demonstrated that LLMs can provide valuable suggestions and identify patterns in complex ICU patient data, but they possess limitations that underscore the continued importance of human expertise in healthcare. One such limitation is the lack of access to comprehensive patient histories and records. This constraint prevents LLMs from fully understanding the context of a patient's condition, which is crucial for making accurate and relevant recommendations.

Moreover, LLMs struggle to comprehend complex clinical information or produce nuanced and detailed clinical descriptions \cite{vaishya2023chatgpt}. These models are primarily trained on textual data and may not have sufficient exposure to specialized medical terminology, complex concepts, or rare conditions, which are vital for effective decision-making in healthcare \cite{vaishya2023chatgpt,sallam2023chatgpt}.

Another area where LLMs face challenges (to some degree) is in interpreting time series data \cite{bosselmann2023ai,xie2023wall} and numerical data \cite{frieder2023mathematical}. The dynamic nature of healthcare particularly the ICU, where clinical status can change in minutes, often requires an understanding of changes in a patient's condition over time and the ability to analyze numerical information, such as laboratory test results and vital signs. LLMs currently lack the sophistication to fully grasp these aspects of patient care.

The findings of our study suggest that there is no inherent contradiction between embracing AI technologies and preserving the roles of healthcare professionals. Instead, LLMs can be seen as a valuable complement to human expertise, providing support and suggestions while healthcare professionals retain ultimate responsibility for patient care. By combining the strengths of AI with the insights and experience of medical professionals, we can work towards a more efficient and effective healthcare system that benefits patients, providers, and society.

\subsection{Improving LLMs for pharmacy}
Optimizing PharmacyGPT for clinical scenarios will require further engineering. First, training LLMs on domain-specific patient data \cite{gu2021domain,liu2022survey} is an established strategy for enhancing their performance in specialized domains. By exposing the model to a larger amount of pharmacy-related data, it will acquire a deeper understanding of the domain's nuances, terminology, and knowledge. This targeted training can lead to improvements in the model's ability to generate relevant, accurate, and contextually appropriate responses for various pharmacy tasks.

Second, refining the model architecture to better align with the specific tasks and objectives in pharmacy can also lead to improvements. By designing customized architectures that cater to the unique requirements of pharmacy applications, LLMs can become more effective in addressing domain-specific challenges \cite{liu2022survey}. This process may involve the development of task-specific layers, attention mechanisms, or other architectural components \cite{rezayi2022agribert,liu2022survey} that are tailored to the needs of pharmacy-related tasks.

Moreover, localizing LLMs to adhere to Health Insurance Portability and Accountability Act (HIPAA) patient guidelines \cite{liu2023deid} is crucial for processing sensitive patient data in hospitals. Ensuring that LLMs are compliant with privacy and security regulations is a vital aspect of their adoption in clinical settings \cite{silvestri2022special}. One approach to achieve this is to develop local, on-premises versions of LLMs that can be integrated into hospital systems, thereby eliminating the need to transmit sensitive data over the internet. Alternatively, techniques such as federated learning \cite{nguyen2022federated} or secure multi-party computation \cite{li2020privacy} can be employed to train LLMs on sensitive data while maintaining patient privacy.

\subsection{Building AI-friendly datasets in pharmacy}
Our results indicate a strong need to create LLM-friendly datasets in pharmacy. 

The foremost approach is to collect more detailed text-based patient information. This process may include extensive documentation of patient histories, treatment plans, and progress notes, which would provide LLMs with a wealth of contextual information for generating accurate and relevant responses. Additionally, incorporating data on patient symptoms, along with standardized clinical assessment scores such as APACHE II \cite{larvin1989apache}, SOFA \cite{lambden2019sofa}, and MRC-ICU \cite{newsome2020medication}, would offer LLMs a more comprehensive understanding of patient conditions, enabling them to make better-informed predictions and recommendations.

Extending the duration of data collection beyond traditional 24-hour or 72-hour windows is another important consideration. By capturing a more extensive timeline of patient information, LLMs can gain deeper insights into the progression of diseases and the effectiveness of different treatment strategies. This longitudinal data would allow LLMs to identify trends, assess long-term outcomes, and provide more accurate predictions for patient recovery and risk factors.

Furthermore, standardizing drug descriptions, including dosage numbers and frequencies, is crucial for creating an LLM-friendly dataset in pharmacy. This standardization can help eliminate ambiguities and inconsistencies in the data, ensuring that LLMs can accurately interpret and generate relevant information about medications. Moreover, the inclusion of drug interactions, contraindications, and side effects can further enhance the LLM's ability to provide safe and effective medication recommendations. The first steps towards standardizing ICU medication data for AI have begun \cite{sikora2023pharmacophenotype}.

Incorporating diverse data sources (e.g., electronic health records, clinical trial results, and relevant scientific literature), can provide LLMs with a more comprehensive understanding of the pharmacy domain. By integrating these diverse sources of information, LLMs can develop a more robust knowledge base, enabling them to generate better-informed responses and recommendations for various pharmacy tasks.

\subsection{Define appropriate NLP tasks and metrics for pharmacy}
To effectively apply LLMs to the pharmacy domain, it is essential to define relevant NLP tasks and develop suitable evaluation metrics. Common NLP tasks include question answering (QA) and summarization \cite{liu2023summary}, while specialized tasks might involve AI-generated prescription plans and outcome predictions. Traditional NLP evaluation metrics, such as n-gram-based ROUGE scores \cite{lin2004rouge} and METEOR scores \cite{banerjee2005meteor}, may not be reasonably applied to these specialized tasks, as they prioritize similarity between generated and reference texts rather than accuracy, safety, and adherence to clinical guidelines.

To address this challenge, domain experts should collaborate to develop specialized evaluation metrics that capture the unique aspects of pharmacy-related tasks. For instance, metrics could assess the correctness of AI-generated prescriptions or the alignment of predicted outcomes with actual patient outcomes, considering sensitivity, specificity, and overall accuracy.

Defining appropriate NLP tasks and developing tailored evaluation metrics for pharmacy applications is crucial for the successful integration of LLMs in this domain. Collaboration between NLP researchers and pharmacy experts is essential to ensure that these models can effectively contribute to improved patient care and clinical practice efficiency.

Because of the complexity and individual nature of ICU pharmacy regimens, which require individualized decision-making, critical care pharmacists, who are experts in this domain, were involved in the evaluation process. Ultimately, critical care pharmacists need to review the medication plans generated by GPT-4 to assess their reasonableness and clinical applicability. This will help ensure that the AI-generated plans align with current best practices and account for the specific needs of each patient. It is also crucial to devise evaluation metrics that extend beyond traditional n-gram-based metrics, such as ROUGE, which may not be sufficient for capturing the nuances of this new task. Developing tailored evaluation metrics for assessing the performance of AI-generated medication plans will enable a more comprehensive understanding of the models' capabilities and limitations. This, in turn, will facilitate improvements in the models and promote the responsible integration of AI in the field of clinical pharmacy. Overall, the outcome and medication prediction results demonstrate a promising approach to using large language models for rapidly assessing patient situations and providing further guidance on medication and treatment plans.

\subsection{Developing Multimodal Foundation Models for Pharmacy}
\label{sec:multimodal}

As the field of artificial intelligence has evolved, the concept of multimodal learning, encompassing the ability to process and integrate data from various sources such as text, images, audio, and structured data, has gained substantial attention \cite{li2023artificial}. This approach is of great value in the pharmacy domain, given the wide variety of data types available, from textual patient notes to numerical lab results, diagnostic images, and structured demographic or medication data. 

Transformer-based architectures, like those found in large language models (LLMs) such as GPT-4, are well-suited to handle diverse types of sequence data and learn intricate patterns within them \cite{liu2022survey,li2023artificial}. These capabilities make them an excellent foundation for developing multimodal models in healthcare and pharmacy. Transformers have proven performance in NLP \cite{zhao2023brain} and have demonstrated success in domains traditionally dominated by convolutional neural network (CNN) based models, including computer vision \cite{wang2023review,dai2023hierarchical,bi2023community} and medical image applications \cite{wang2023review,zhang2023differentiating,zhang2023beam,ding2023deep,dai2022graph,ding2022accurate,qiang4309357deep,liu2022discovering}. Additionally, they have been utilized in other modalities such as modeling human brain functions \cite{zhao2023generic,zhao2022embedding}. For example, it powers vision foundation models such as the Segment Anything Model (SAM) \cite{kirillov2023segment,zhang2023segment,dai2023samaug}. This demonstrates their flexibility and potential in handling multiple data modalities \cite{li2023artificial}.

The multimodal models for pharmacy would involve creating distinct input pathways for each type of data, each leveraging the self-attention mechanisms of transformers. After processing each modality independently, the outputs can be integrated to generate a comprehensive patient representation. This fusion can be guided by various strategies and can determine which features from each modality are most relevant for a given task.

When trained on a diverse dataset of patients, a multimodal foundation model could potentially generate more accurate and personalized medication plans by considering the full range of patient data. In this way, the model can provide an in-depth understanding of the patient's current health status, medication needs, and potential risk factors, leading to more effective and tailored pharmaceutical care.

\section{Conclusion}
In conclusion, our exploration into the application of large language models in pharmacy, embodied by PharmacyGPT, illuminates promising avenues for future development. Despite the evident challenges, we believe that leveraging LLMs can drastically enhance the accuracy, personalization, and efficiency of medication plan generation. The potential for improved patient outcomes and streamlined pharmaceutical operations is significant. As we continue to refine these models, a keen focus on the incorporation of expert knowledge, carefully designed training datasets, and development of tailored metrics for model evaluation will be crucial. By continuing to intertwine AI with pharmacy, we stand at the precipice of a new era of healthcare where AI augments the crucial human component, leading to comprehensive, high-quality patient care.




\bibliography{LLM_refs}
\bibliographystyle{unsrt}

\end{document}